\documentclass{article}[12pt]

\PassOptionsToPackage{numbers, compress}{natbib}
\usepackage[preprint]{nips_2018}
\usepackage[utf8]{inputenc} 
\usepackage[T1]{fontenc}    
\usepackage{hyperref}       
\usepackage{url}            
\usepackage{booktabs}       
\usepackage{amsfonts}       
\usepackage{nicefrac}       
\usepackage{microtype}      

\usepackage{color}
\usepackage{enumitem}
\usepackage{graphicx}
\newlist{step}{enumerate}{1}
\setlist[step]{label=Step~\arabic*:}

\date{18th May 2018}
\begin{document}
\date{\currenttime}

\title{Cavity Filling: Pseudo-Feature Generation for Multi-Class Imbalanced Data Problems in Deep Learning}

\author{Tomohiko Konno\thanks{Corresponding Author: \color{blue}{tomohiko@nict.go.jp}}~~ and ~Michiaki Iwazume \\
National Institute of Information and Communications Technology, Tokyo Japan}
\maketitle

\begin{abstract}
Herein, we generate pseudo-features based on the multivariate probability distributions obtained from the feature maps in layers of trained deep neural networks. Further, we augment the minor-class data based on these generated pseudo-features to overcome the imbalanced data problems. 
The proposed method, i.e., cavity filling, improves the deep learning capabilities in several problems because all the real-world data are observed to be imbalanced. 
\end{abstract}

\section{Introduction: Imbalanced Data Problems}
If some classes in a dataset contain few samples, the accuracy and other characteristics of these classes considerably decrease, which can cause problems during the application of machine leaning. When 90\% of the data are negative and the remaining 10\% are positive, an accuracy of 90\% can be obtained if all the data are assumed to be negative. Therefore, positive samples are not detected. The machine learning algorithms tend to learn in a similar manner in case of imbalanced data because learning is driven by major classes containing large volumes of data. Further, a minor class contains a comparatively small volume of data.

\textbf{Examples of real imbalanced data problems} observed in reality include 
bioinfomatics~\cite{garcia2012class}, security~\cite{cieslak2006combating,dal2014learned,tesfahun2013intrusion,phua2004minority}, finance~\cite{sanz2015compact}, satellite imaging~\cite{zheng2015support}, medicine~\cite{eskildsen2014detecting,krawczyk2016evolutionary}, software development~\cite{garcia2009evolutionary,mollineda2007class}, fault diagnosis~\cite{zhang2015deep}, risk management~\cite{ezawa1996learning}, brain computer interface~\cite{goh2014artifact}, medical diagnosis~\cite{valdovinos2005class,grzymala2004approach}, tool condition monitoring~\cite{sun2004multiclassification}, activity recognition~\cite{gao2016adaptive}, video mining~\cite{gao2014enhanced}, sentiment analysis~\cite{munkhdalai2015self}, behavior analysis~\cite{azaria2014behavioral}, text mining~\cite{munkhdalai2015self}, industrial system monitoring~\cite{ramentol2016fuzzy}, target detection~\cite{razakarivony2016vehicle}, software defect prediction~\cite{saez2015smote}, hyperspectral data analysis~\cite{sun2009classification}, and disease detection~\cite{2018arXiv180502730W}.

\subsection{Existing Solutions and Background}
The imbalanced data problems were reviewed in previously conducted studies~\cite{he2009learning, galar2012review,Krawczyk2016}; in this subsection, typical methods for solving the imbalanced data problems will be reviewed. In the following examples, let us assume that the data contain major and minor classes with 5,000 and 500 samples each, respectively. The concept of these methods is to balance the data among various classes. \textbf{Oversampling} resamples minor-class data to balance them. In this example, minor classes are resampled 10 times. However, random oversampling may result in data overfitting. In contrast, \textbf{undersampling} reduces the major-class data for balancing the dataset. In this case, only 500 samples would be used in each major class. However, random undersampling may eliminate important data. Random elimination of the major-class data is not the only undersampling method. For instance, \textbf{informed sampling}, where the sampling weight is calculated, can leave important data. In this example, 4,500 of the 5,000 samples in each major class are disposed, which denotes a wastage, especially because the deep learning algorithms require large volumes of data. Thus, undersampling does not occasionally work well in deep learning. As we can observe from our experiments, undersampling improves the minor class performance; however, the total amount of data decreases, reducing the accuracy, precision, recall, and f1 scores, which may cause other problems. 

\begin{figure}[htbp]
\begin{center}
    \includegraphics[clip,width=0.7\hsize]{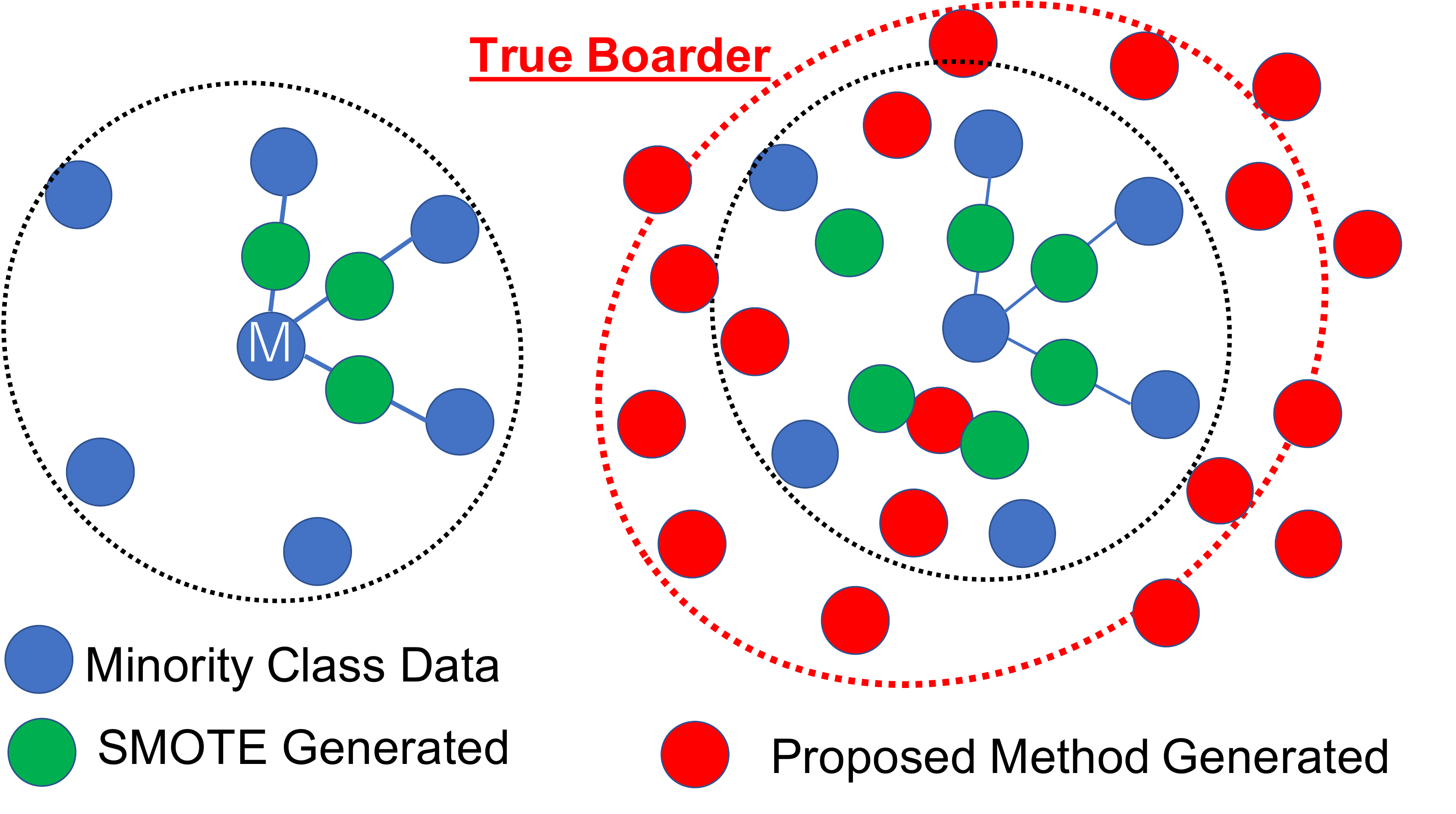}
    \caption{Left: In SMOTE, the minor-class data are randomly selected (denoted as M in the figure). Further, k-neighbors (3-neighbors in this study) are selected, and pseudo-data are generated at the interpolating points. Right: Even though the pseudo-data generated by SMOTE cannot go beyond the determination border, our proposed method can go beyond the border and push it forward near to the true determination border.}\label{fig: smote}
    \end{center}
\end{figure}

The synthetic minor class oversampling technique (\textbf{SMOTE})~\cite{chawla2002smote} is a pseudo-data generation method. In SMOTE, the minor-class data k-neighbors are selected, and pseudo-data are generated at the interpolations of k-neighbors, as illustrated on the left side of Fig.~\ref{fig: smote}. Theoretically, pseudo-data cannot cross the original determination borders in SMOTE, whereas our proposed method, i.e., cavity filling, can cross the border and push it forward, as illustrated on the right side of Fig.~\ref{fig: smote}.

In deep learning feature spaces, SMOTE is used to augment data, though not for imbalanced data problems, and has been previously studied~\cite{2017arXiv170205538D}. Further, we compare SMOTE in feature spaces with cavity filling and denote that our proposed method always outperforms SMOTE in the experiments.

The \textbf{weight-adjusted loss function} exists in addition to the sampling methods. A weight-adjusted loss function aims to reduce the weights of the major classes and increase those of the minor classes in the loss function. In our example, the minor-class losses increase by $10$ times when compared with the major-class losses.

\textbf{Bagging and boosting}: Many methods are available to reduce the major-class data from 5,000 to 500 in undersampling. Further, the corresponding classifiers can be constructed, and ensemble learning can be achieved using these classifiers. Ensemble learning with undersampling is a good methodology~\cite{wallace2011class}; however, ensemble learning can be computationally expensive in case of deep learning.

\textbf{Combination of methods}: In real applications, we solve imbalanced data using various combinations of methods and do not use only a single method. Therefore, we add one method that is worth testing to other methods.

\subsubsection{Background: Recent Progress and Success in Deep Learning}
As will be clarified by the remainder of this study, the success of our proposed method can be attributed to the achievements in deep learning. Deep neural networks accurately capture the feature distributions; thus, multidimensional probability distributions that accurately fit the features can be obtained. \emph{It was not until recent progress and success in deep learning that our proposed method would have been possible. Thus, our proposed method belongs to the deep learning age.}

\subsection{Contributions}
The contributions of our study can be given as follows.
\begin{enumerate}
    \item We propose a method named ``cavity filling'' to synthesize pseudo-features based on the multivariate probability distributions obtained from feature maps in a layer of trained deep neural networks.
    \item We virtually increase the data volume in minor classes using the generated pseudo-feature and make progress in multi-class imbalanced data problems in deep learning.
\end{enumerate}
In cavity filling, we use the existing networks, such as ResNet~\cite{he2015deep}, in the experiments, and \textbf{we do not have to modify the successful network structures, which is an advantage of our proposed method.} Changing the structures of the existing successful networks may cause decreased performance.

\subsection{Recent Works}
Recent works on imbalanced data can be summarized as follows. The methods for imbalanced data in convolutional neural networks were compared\cite{2017arXiv171005381B}, and instance selections and geometric mean accuracy were studied~\cite{2018arXiv180407155K}. Further, the cost-sensitive deep belief networks were studied~\cite{2018arXiv180410801Z}, and a cost function approach, i.e., class rectification loss, was studied~\cite{2017arXiv171203162D}. The neighbors progressive competitive algorithm, inspired by the k-nearest neighbor, has also been proposed~\cite{2017arXiv171110934S}. A boosting method known as locality informed under-boosting was proposed in a previous study~\cite{2017arXiv171105365A}. Furthermore, an imbalanced cardiovascular medical dataset was previously studied~\cite{Rahman2013AddressingTC}.Supplemental data selection for data rebalancing was also studied~\cite{2018arXiv180102548M}. An approach to train the generative adversarial networks using imbalanced data and to generate the minor-class images was studied in~\cite{2018arXiv180309655M}, and an example reweighing algorithm for deep learning was investigated~\cite{2018arXiv180309050R}. A sampling method with a loss function, i.e., quintuplet sampling with triple-header loss, was also previously investigated~\cite{huang2016learning}.

\section{The Proposed Method: Cavity Filling}\label{sec: method}
\begin{figure}[htbp]
\begin{center}
    \includegraphics[width=0.7\hsize]{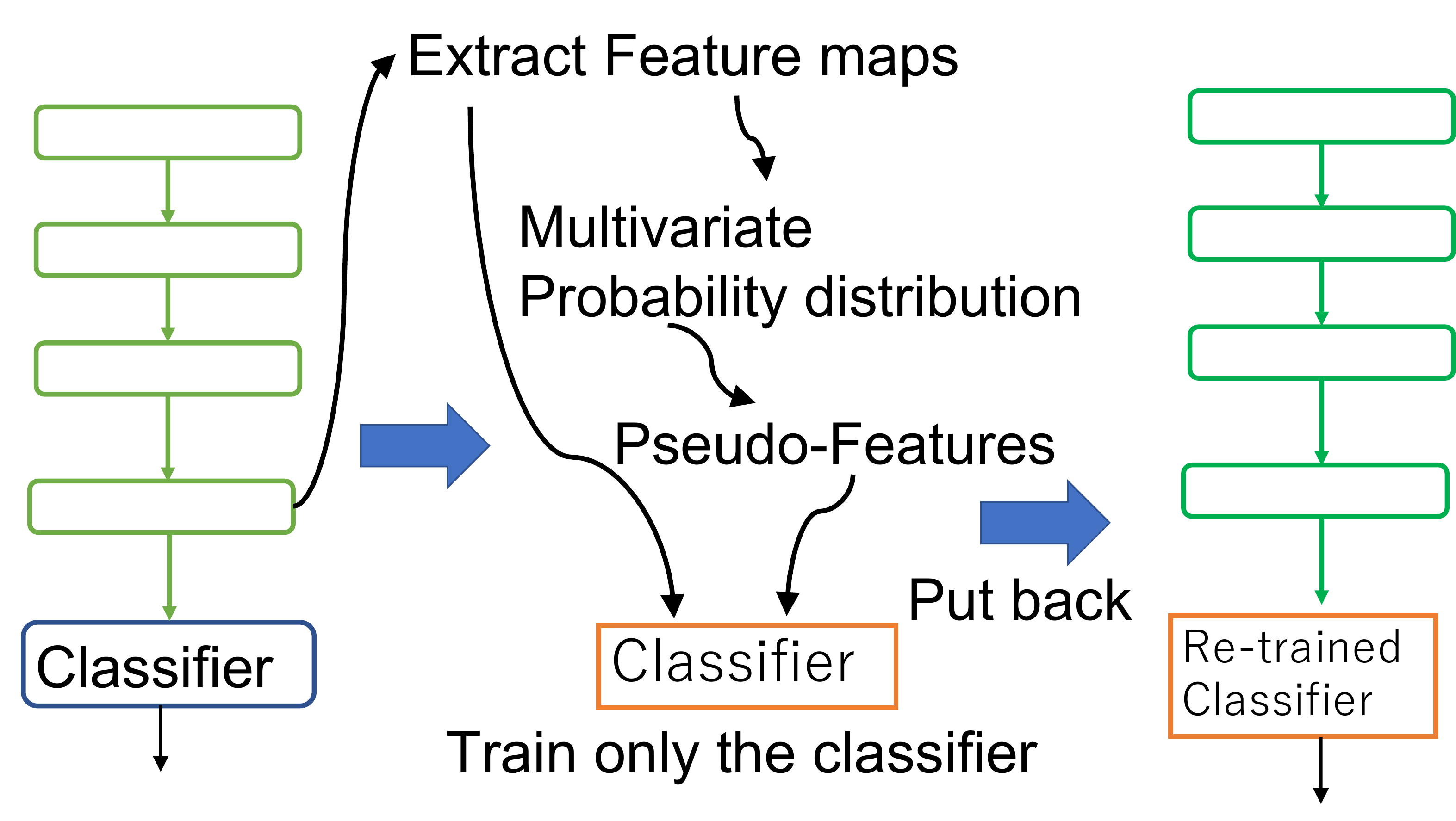}
    \caption{A schematic of the proposed cavity filling method. Left: Train deep neural networks. Center: Extract features from a layer, obtain multivariate probability distributions of these features, and generate pseudo-features of minor classes from the probability distributions and retrain the following layers. Right: Return the retrained layers to the original one (the final classifier is retrained and substituted into the experiment.)}\label{fig: network_1}
    \end{center}
\end{figure}
\begin{figure}[htbp]
\begin{center}
    \includegraphics[width=0.6\hsize]{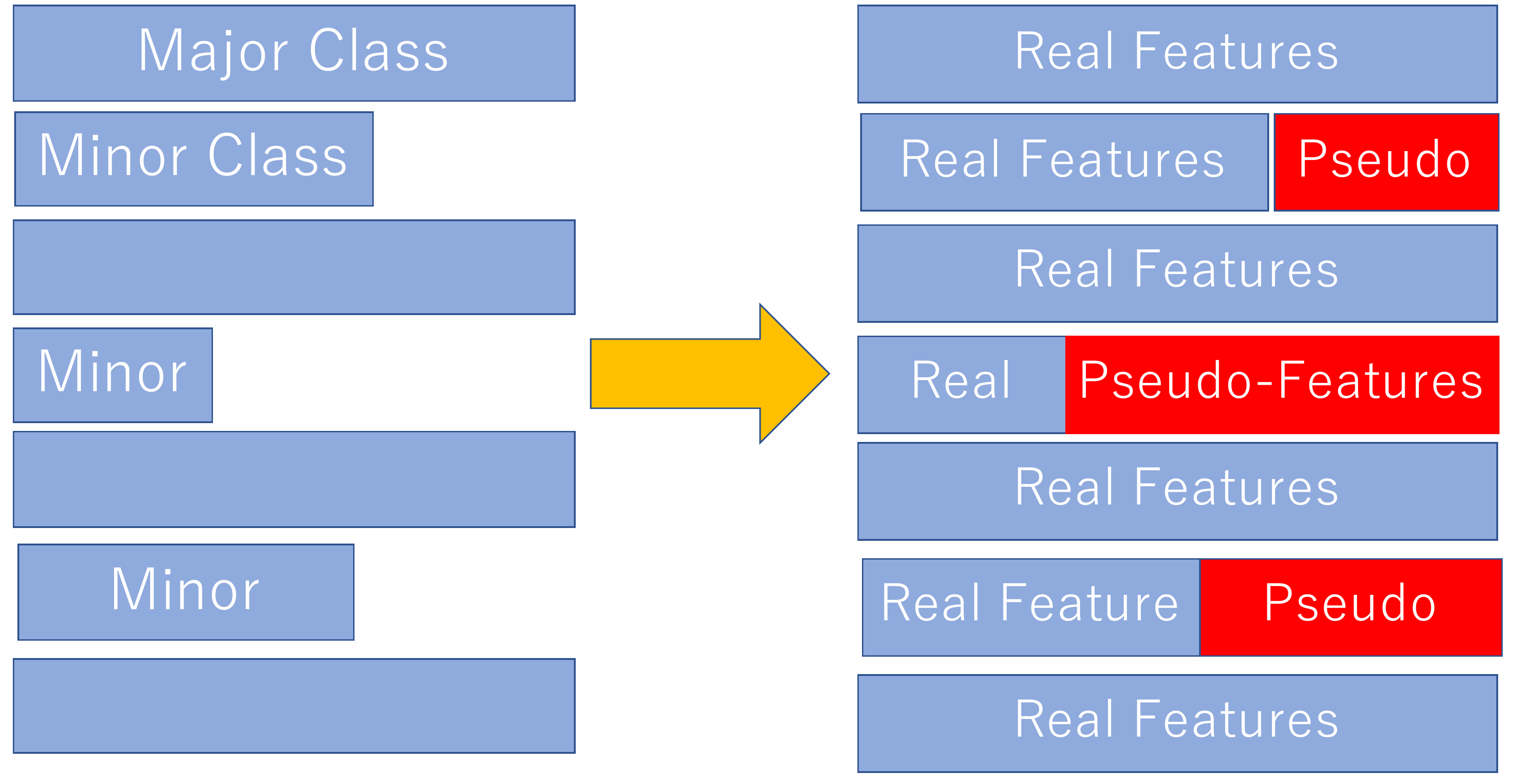}
    \caption{The cavities of the original imbalanced data (left) are filled by pseudo-features in the feature spaces (right). The pseudo-features are generated using the multivariate probability distributions obtained from the real features obtained by deep learning.}\label{fig: cavity-filling}
    \end{center}
\end{figure}
Our proposed method can be explained as follows. We require two-stage training. Fig.~\ref{fig: network_1} and Fig.~\ref{fig: cavity-filling} depict the diagrams of the procedure and its main idea, respectively.
\begin{step}
    \item Train deep neural networks.
    \item Extract feature maps from a layer.
    \item Obtain multivariate probability distributions of these feature maps.
    \item Generate pseudo-features from the multivariate probability distributions to obtain pseudo-features of the minor classes.
    \item Train the layers following the layer from which features were extracted based on the real features and pseudo-features. Only the classifier layers are updated.
    \item Return the trained classifier layers to the original network and use the newly combined network for performing the estimation.
\end{step}

In the experiments, deep neural networks are trained using the original imbalanced data. Further, features are extracted from the layer immediately before the classifier layer and a multivariate Gaussian method is used to parameterize the features. We observed that the multivariate Gaussian method works better than an independent Gaussian method comprising independent random variables assigned to each feature-map dimension in the experiments. The final classifier is retrained based on the real features and pseudo-features, and the retrained classifier is returned to the network.




\section{Experiments on Multi-Class Imbalanced Data Problem}
We synthesized an imbalanced dataset using Cifar10~\cite{krizhevsky2009learning} and Imagenet~\cite{ILSVRC15} datasets. Subsequently, we conducted two experiments using the imbalanced dataset.



We assume that we need some amount of data, even for minor classes, because cavity filling considerably relies on the deep-learning captured features. However, in the deep learning age, private companies possess large datasets with large volumes of data, even for minor classes. In this situation, cavity filling works appropriately. 


\subsection{Synthesizing the Multi-Class Imbalanced Data (Cifar10)}
We synthesized imbalanced data from cifar10~\cite{krizhevsky2009learning} because it is a standard benchmark dataset. In cifar10, $10$ classes are present with $5,000$ samples in each class. Thus, the total number of samples in the training data and test data is $50,000$ and $10,000$, respectively. Subsequently, we determine the number of classes that can be considered minor classes and randomly select these classes; the remaining classes become major classes. We then reduce the data volume in minor classes by a factor of $10$. Thus, minor classes contain $500$ samples, whereas major classes contain $5,000$ samples in our experiments\footnote{if we consider minor classes in a similar manner using cifar100, each minor class would contain only $50$ samples, which is not suited for deep learning.}.

\subsection{Comparison of Methods (Cifar10)}
The baseline denotes training using the original imbalanced data. In the experiments, undersampling indicates that the major-class data are randomly reduced from $5,000$ to $500$ to balance the classes. Thus, oversampling denotes that the $500$ minor-class data are resampled $5,000$ times. SMOTE\footnote{The python package "imbalanced-learn"~\cite{JMLR:v18:16-365} is used.} contains training of the classifier based on the real features and added pseudo-features made by SMOTE in the feature space. Perturbed denotes training of the classifier with additional perturbed features of multidimensional Gaussian noise, which has a mean of $0$ and a variance calculated based on the real features of the training data. Finally, proposed denotes the proposed method, i.e., cavity filling. Pseudo-features are generated in SMOTE, perturbed, and proposed; thus, the real and pseudo-features constitute balanced data. More specifically, $4500$ pseudo-features are produced for each minor class. 

\subsection{Experimental Flow (Cifar10)}
We changed the number of minor classes from $1$ to $9$ and performed experiments as follows.
\begin{enumerate}
    \item Select the number of minor classes, varying from $1$ to $9$ throughout the experiment.
    \item Use $500$ from among $5,000$ samples in minor classes, producing imbalanced data from cifar10. The imbalance ratio is $1$ to $10$.
    \item By keeping the minor classes fixed, the baseline, undersampling, oversampling, SMOTE, perturbed, and cavity filling methods are compared.
\end{enumerate}
Keras~\cite{chollet2015keras} and ResNet56~\cite{he2015deep} are used, with a batch size of $128$, an optimizer of Adam, and epochs of $100$. The test data are original cifar10 test data.


%

\begin{figure}[htbp]
\begin{center}
    \includegraphics[clip,width=\hsize]{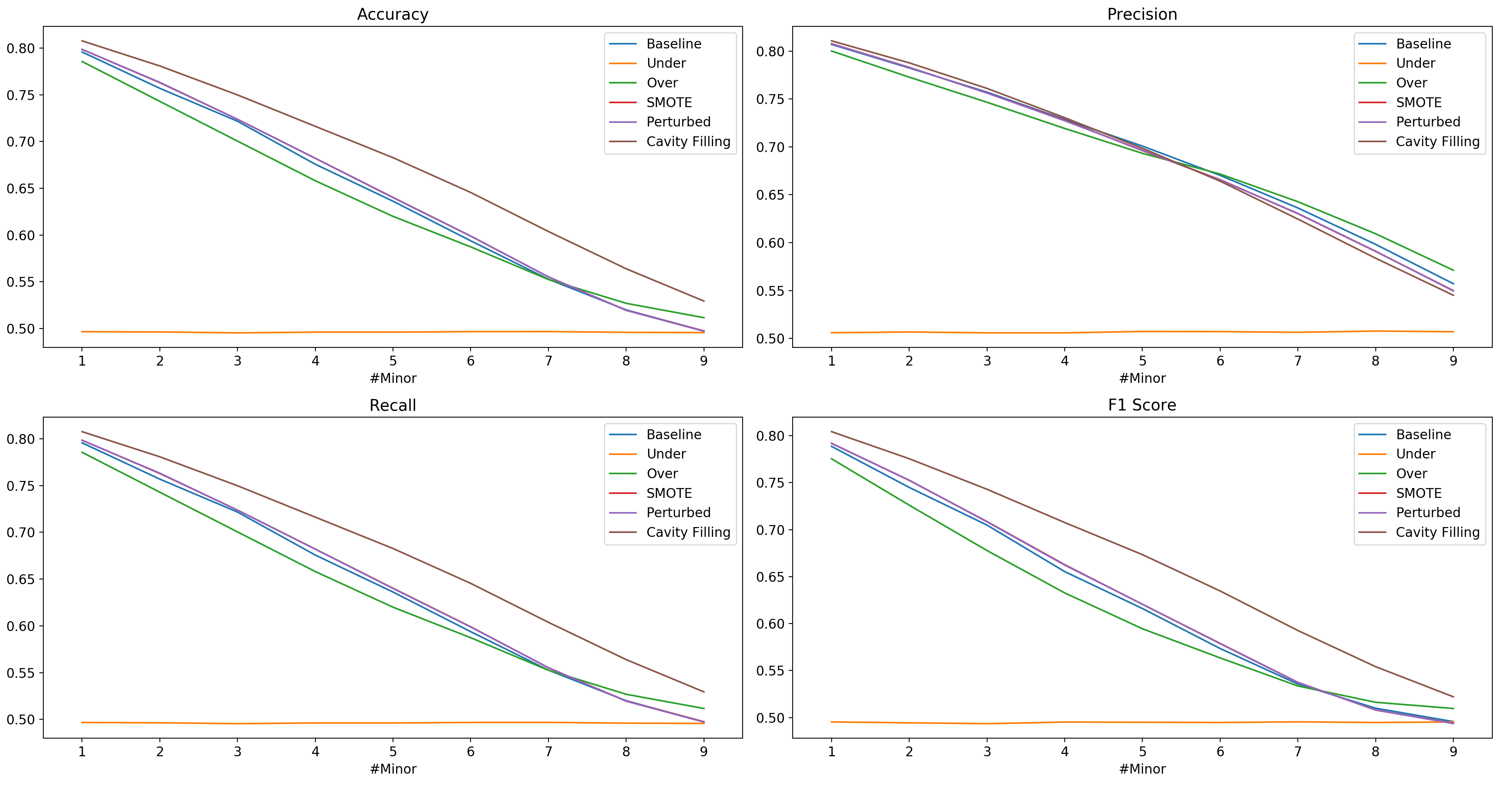}
    \caption{Comparison of six methods for handling the multi-class imbalanced data (Cifar10): \#Minor denotes the number of minor classes. Top left: accuracy, bottom left: recall, top right: precision, and bottom right: F1}\label{fig: figure4}
    \end{center}
%
\begin{center}
    \includegraphics[clip,width=\hsize]{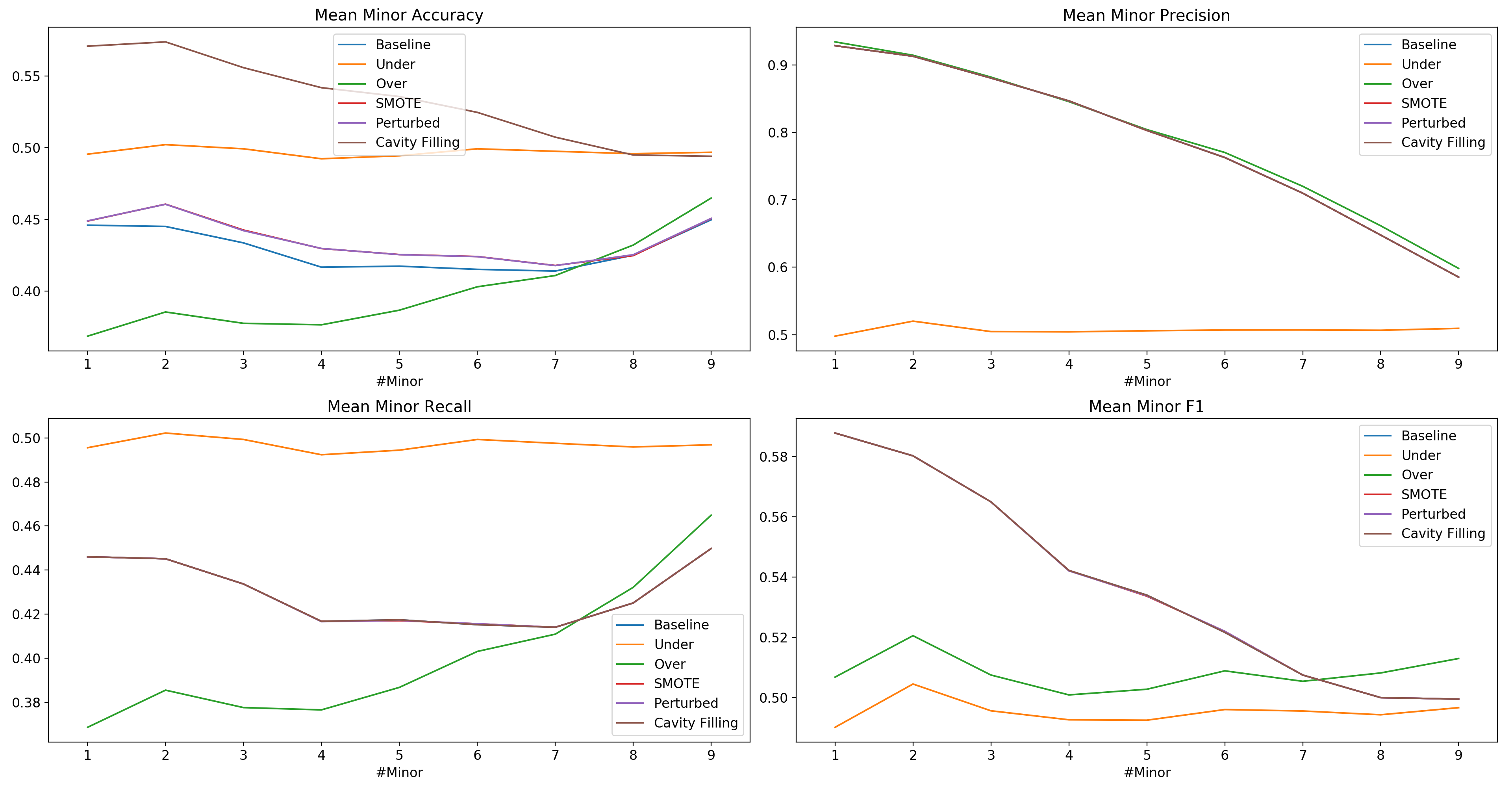}
    \caption{Comparison of the accuracy, precision, recall, and F1 in only minor classes. Undersampling, SMOTE, perturbation, and cavity filling exhibit identical scores and overlap; thus, they cannot be distinguished.}\label{fig: figure5}
    \end{center}
\end{figure}

\subsection{Experimental Results (Cifar10)}
The experimental results are the averages obtained over 144 episodes. One episode indicates a sequence of experiments in which the number of minor classes changes from $1$ to $9$. \#Minor denotes the number of minor classes and does not represent a class identification number. If \#Minor is $4$, $4$ out of $10$ classes contain only $500$ samples, whereas the remaining classes contain $5,000$ samples. Once the minor classes were determined, we performed baseline, undersampling, oversampling, SMOTE, perturbed, and cavity filling methods while keeping the minor classes fixed. 

The results are illustrated in Fig.~\ref{fig: figure4} and Fig.~\ref{fig: figure5}. 
In Fig.~\ref{fig: figure4}, the accuracy is for the whole data, whereas the precision, recall, and F1 are macro averages for each class. The proposed method is much better in terms of accuracy, recall, and F1 and is as good as other methods with respect to the precision.

We also verified the mean accuracy, precision, recall, and F1 of the minor classes alone ,and they are the macro averages over minor classes because the scores in minor classes are important at times. They are illustrated in Fig.~\ref{fig: figure5}. Undersampling, SMOTE, perturbation, and cavity filling exhibit identical mean precision, recall, and F1 scores of minor classes and cannot be distinguished in the figures. The scores are summarized in Tab.~\ref{tab: acc}--Tab.~\ref{tab: mean_min_f1} in the appendix. Although undersampling exhibits improved mean minor recall, cavity filling is better than or as good as undersampling and oversampling with respect to the mean minor accuracy, precision, and F1. Further, cavity filling and undersampling can be combined. We can use undersampling and cavity filling for other cases when the recall of minor classes is important.

\subsection{Experiment on Multi-Class Imbalanced Data Obtained from ImageNet}
We synthesized imbalanced data from ImageNet, comprising $1,000$ image classes and approximately $13$M images.
Further, we produced imbalanced data from ImageNet as follows. First, we chose $500$ minor classes among $1,000$ classes. Then, we reduced the number of images in minor classes by a factor of $10$, yielding an imbalance ratio of $10$ to $1$.

\begin{table}
\centering
\caption{Scores for imbalanced data produced from ImageNet}\label{tab: imagenet}
\begin{tabular}{|c|c|c|c|c|}
\hline
 & Accuracy & Precision & Recall & F1 \\ \hline
baseline & 49.9  &  51.5  &  43.6  &  40.0  \\ \hline
undersampling&29.7&29.0&29.5&29.0\\ \hline
over sampling & 49.5  &  51.6  &  43.9  &  40.7  \\ \hline
SMOTE & 53.5  &  55.2  &  53.4  &  52.8 \\ \hline
\textbf{Cavity Filling} & \textbf{54.2}  & \textbf{ 55.0}  & \textbf{ 54.1}  & \textbf{53.5 } \\ \hline
\end{tabular}
\end{table}

We used the Resnet34 neural network, an optimizer of Adam, no data augmentation, a batch size of $256$, and an epoch of $100$. The baseline values in the table are for the as-is imbalanced data and accuracy for the original test data. The precision, recall, F1 are macro averages over each class; therefore, they are low. However, they are improved by cavity filling. The results are summarized in Table~\ref{tab: imagenet}. Cavity filling improves the accuracy by $4.3\%$ from baseline, making it better than SMOTE.


\section{Conclusion}
We propose a method named cavity filling, which generates pseudo-features to fill the gaps between the minor and major classes in feature spaces. Further, we extract features from a layer of trained deep neural network, obtain multivariate probability distributions from the features of each minor class, and sample the minor-class pseudo-features from multivariate probability to virtually increase the minor-class training data. We do not generate pseudo-data but generate pseudo-features. Subsequently, we train the layers following the layer from which features were extracted based on the pseudo-features and real features. The cavity filling method functions in an appropriate manner for multi-class imbalanced data.

\bibliographystyle{unsrt}
\bibliography{bib_pseudo_data_generation.bib}
\textbf{Competing Interests:} The authors declare that they have no
 competing financial interests.\\
\textbf{Correspondence:}
Correspondence and requests for materials should be addressed to Tomohiko Konno (tomohiko@nict.go.jp).

\newpage

\appendix
\section{Mean scores of the minor classes (Cifar10)}
The tables that summarize the macro averaged scores for minor classes in Cifar10 are presented here because some points are difficult to distinguish in the figures.
Under represents undersampling, Over represents oversampling, and Perturbed represents perturbation.

\begin{table}[htbp]
    \begin{center}
        \begin{tabular}{|c|c||c|c|c|c||c|}\hline
 \#Minor&Baseline &  Under &  Over &  SMOTE &  Perturbed &  Cavity Filling\\ \hline
  1&   0.45 &   0.50 &  0.37 &   0.45 &       0.45 &      0.57 \\ \hline
 2 &   0.45 &   0.50 &  0.39 &   0.46 &       0.46 &      0.57 \\ \hline
   3 & 0.43 &   0.50 &  0.38 &   0.44 &       0.44 &      0.56 \\ \hline
     4&0.42 &   0.49 &  0.38 &   0.43 &       0.43 &      0.54 \\ \hline
     5&0.42 &   0.49 &  0.39 &   0.43 &       0.43 &      0.54 \\ \hline
6     &0.42 &   0.50 &  0.40 &   0.42 &       0.42 &      0.52 \\ \hline
  7   &0.41 &   0.50 &  0.41 &   0.42 &       0.42 &      0.51 \\ \hline
    8 &0.43 &   0.50 &  0.43 &   0.42 &       0.43 &      0.50 \\ \hline
     9&0.45 &   0.50 &  0.46 &   0.45 &       0.45 &      0.49 \\ \hline
\end{tabular}
\caption{Mean Minor Accuracy}\label{tab: acc}
    \end{center}
    \begin{center}
        \begin{tabular}{|c|c||c|c|c|c||c|}\hline
\#Minor& Baseline &  Under &  Over &  SMOTE &  Perturbed &  Cavity Filling \\ \hline
1     &0.93 &   0.50 &  0.93 &   0.93 &       0.93 &      0.93 \\ \hline
 2    &0.91 &   0.52 &  0.91 &   0.91 &       0.91 &      0.91 \\ \hline
  3&   0.88 &   0.50 &  0.88 &   0.88 &       0.88 &      0.88 \\ \hline
   4&  0.85 &   0.50 &  0.85 &   0.85 &       0.85 &      0.85 \\ \hline
    5& 0.80 &   0.51 &  0.80 &   0.80 &       0.80 &      0.80 \\ \hline
    6& 0.76 &   0.51 &  0.77 &   0.76 &       0.76 &      0.76 \\ \hline
    7& 0.71 &   0.51 &  0.72 &   0.71 &       0.71 &      0.71 \\ \hline
    8& 0.65 &   0.51 &  0.66 &   0.65 &       0.65 &      0.65 \\ \hline
     9&0.59 &   0.51 &  0.60 &   0.59 &       0.59 &      0.59 \\ \hline 
\end{tabular}\caption{Mean Minor Precision} \label{tab: mean_min_prec}
    \end{center}
    \begin{center}
        \begin{tabular}{|c|c||c|c|c|c||c|}\hline
\#Minor&Baseline &  Under &  Over &  SMOTE &  Perturbed &  Cavity Filling \\ \hline
1&     0.45 &   0.50 &  0.37 &   0.45 &       0.45 &      0.45 \\ \hline
 2 &   0.45 &   0.50 &  0.39 &   0.45 &       0.45 &      0.45 \\ \hline
  3 &  0.43 &   0.50 &  0.38 &   0.43 &       0.43 &      0.43 \\ \hline
    4& 0.42 &   0.49 &  0.38 &   0.42 &       0.42 &      0.42 \\ \hline
 5  &  0.42 &   0.49 &  0.39 &   0.42 &       0.42 &      0.42 \\ \hline
  6 &  0.42 &   0.50 &  0.40 &   0.42 &       0.42 &      0.42 \\ \hline
   7&  0.41 &   0.50 &  0.41 &   0.41 &       0.41 &      0.41 \\ \hline
    8 &0.43 &   0.50 &  0.43 &   0.43 &       0.43 &      0.43 \\ \hline
    9& 0.45 &   0.50 &  0.46 &   0.45 &       0.45 &      0.45 \\ \hline        
\end{tabular}\caption{Mean Minor Recall}\label{tab: mean_min_rec}
    \end{center}
%
    \begin{center}
        \begin{tabular}{|c|c||c|c|c|c||c|}\hline
 \#Minor& Baseline &  Under &  Over &  SMOTE &  Perturbed &  Cavity Filling \\ \hline
1&     0.59 &   0.49 &  0.51 &   0.59 &       0.59 &      0.59 \\ \hline
 2 &   0.58 &   0.50 &  0.52 &   0.58 &       0.58 &      0.58 \\ \hline
  3  & 0.56 &   0.50 &  0.51 &   0.56 &       0.56 &      0.56 \\ \hline
  4   &0.54 &   0.49 &  0.50 &   0.54 &       0.54 &      0.54 \\ \hline
   5 & 0.53 &   0.49 &  0.50 &   0.53 &       0.53 &      0.53 \\ \hline
   6 & 0.52 &   0.50 &  0.51 &   0.52 &       0.52 &      0.52 \\ \hline
    7& 0.51 &   0.50 &  0.51 &   0.51 &       0.51 &      0.51 \\ \hline
   8 & 0.50 &   0.49 &  0.51 &   0.50 &       0.50 &      0.50 \\ \hline
    9& 0.50 &   0.50 &  0.51 &   0.50 &       0.50 &      0.50 \\ \hline
\end{tabular} \caption{Mean Minor F1}\label{tab: mean_min_f1}
    \end{center}
\end{table}

\end{document}